January 14, 2019

# Data science is science's second chance to get causal inference right. A classification of data science tasks

Miguel A. Hernán,[1,2] John Hsu[3,4], Brian Healy[5,6]

1. Departments of Epidemiology and Biostatistics, Harvard T.H. Chan School of Public Health, Boston, MA
2. Harvard-MIT Division of Health Sciences and Technology, Boston, MA
3. Mongan Institute, Massachusetts General Hospital, Boston, MA
4. Department of Health Care Policy, Harvard Medical School, Boston, MA
5. Department of Neurology, Harvard Medical School, Partners MS Center, Brigham and Women's Hospital, Boston, MA
6. Biostatistics Center, Massachusetts General Hospital, Boston, MA

Correspondence: Miguel Hernán, Department of Epidemiology, Harvard T.H. Chan School of Public Health, 677 Huntington Avenue, Boston, MA 02115; email: miguel_hernan@post.harvard.edu

**Abstract**

Causal inference from observational data is the goal of many data analyses in the health and social sciences. However, academic statistics has often frowned upon data analyses with a causal objective. The introduction of the term "data science" provides a historic opportunity to redefine data analysis in such a way that it naturally accommodates causal inference from observational data.

Like others before, we organize the scientific contributions of data science into three classes of tasks: Description, prediction, and counterfactual prediction (which includes causal inference). An explicit classification of data science tasks is necessary to discuss the data, assumptions, and analytics required to successfully accomplish each task.

We argue that a failure to adequately describe the role of subject-matter expert knowledge in data analysis is a source of widespread misunderstandings about data science. Specifically, causal analyses typically require not only good data and algorithms, but also domain expert knowledge. We discuss the implications for the use of data science to guide decision-making in the real world and to train data scientists.

**Introduction**

For much of science's recent history, learning from data was the academic realm of Statistics.[1,2] But, in the early 20$^{th}$ century, the founders of modern Statistics made a momentous decision about what could and could not be learned from data. They proclaimed that statistics could be applied to make causal inferences when using data from randomized experiments, but not when using nonexperimental (observational) data.[3-5] This decision classified an entire class of scientific questions in the health and social sciences as not amenable to formal quantitative inference.

Not surprisingly, the statisticians' decree was ignored by many scientists, who continued to use observational data to study the unintended harms of medical treatments, the health effects of lifestyle activities or the social impact of educational policies. Unfortunately, these scientists' causal questions often were mismatched with their statistical training. Perplexing paradoxes arose—for example, the famous "Simpson's paradox" stemmed from a failure to recognize that the choice of data analysis depends on the causal structure of the problem.[6] Mistakes occurred. For example, a generation of medical researchers and clinicians believed that postmenopausal hormone therapy reduced the risk of heart disease because of data analyses that deviated from basic causal considerations. Even today, the confusions generated by a century-old refusal to tackle causal questions explicitly are widespread in scientific research.[7]

To bridge science and data analysis, a few rogue statisticians, epidemiologists, econometricians, and computer scientists developed formal methods to quantify causal effects from observational data. Initially, each discipline emphasized different types of causal questions, developed different terminologies, and preferred different data analysis techniques. By the beginning of the 21$^{st}$ century, while some conceptual discrepancies remained,[8,9] a unified theory of quantitative causal inference had emerged.

We now have a historic opportunity to redefine data analysis in such a way that it naturally accommodates a science-wide framework for causal inference from observational data. A recent influx of data analysts, many not formally trained in statistical theory, bring a fresh attitude that does not a priori exclude causal questions. This new wave of data analysts refer to themselves as data scientists and to their activities as data science, a term popularized by technology companies and embraced by academic institutions.

Data science, as an umbrella term for all types of data analysis, can tear down the barriers erected by traditional statistics, put data analysis at the service of all scientific questions, including causal ones, and prevent unnecessary inferential mistakes. But our chance to successfully integrate data analysis into all scientific questions may be missed if data science ends up being defined exclusively around technical[10] activities (management, processing, analysis, visualization…) without explicit consideration of the scientific tasks.

**A classification of data science tasks**

Data scientists often define their work as "gaining insights" or "extracting meaning" from data. These definitions are too vague to characterize the scientific uses of data science. Only by precisely classifying the "insights" and "meaning" that data can provide will we be able to think systematically about the types of data, assumptions, and analytics that are needed. The scientific contributions of data science can be organized into three classes of tasks: description, prediction, and counterfactual prediction (see Table for examples of research questions for each of these tasks).

<u>Description</u> is using data to provide a quantitative summary of certain features of the world. Descriptive tasks include, for example, computing the proportion of individuals with diabetes in a large healthcare database and representing social networks in a community. The analytics employed for description range from elementary calculations (e.g., a mean or a proportion) to sophisticated techniques such as unsupervised learning algorithms (e.g., cluster analysis) and clever data visualizations.

<u>Prediction</u> is using data to map some features of the world (the inputs) to other features of the world (the outputs). Prediction often starts with simple tasks (e.g., quantifying the association between albumin levels at admission and death within one week among patients in the intensive care unit) and then progresses to more complex ones (e.g., using hundreds of variables measured at admission to predict which patients are more likely to die within one week). The analytics employed for prediction range from elementary calculations (e.g., a correlation coefficient or a risk difference) to sophisticated pattern recognition methods and supervised learning algorithms that can be used as classifiers (e.g., random forests, neural networks) or to predict the joint distribution of multiple variables.

Counterfactual prediction is using data to predict certain features of the world if the world had been different, which is required in *causal inference* applications. An example of causal inference is the estimation of the mortality rate that would have been observed if all individuals in a study population had received screening for colorectal cancer vs. if they had not received screening. The analytics employed for causal inference range from elementary calculations in randomized experiments with no loss to follow-up and perfect adherence (e.g., the difference in mortality rates between the screened and the unscreened) to complex implementations of g-methods in observational studies with treatment-confounder feedback (e.g., the plug-in g-formula).[11] Note that, contrary to some computer scientists' belief, "causal inference" and "reinforcement learning" are not synonyms. Reinforcement learning is a technique that, in some simple settings, leads to sound causal inference. However, reinforcement learning is insufficient for causal inference in complex causal settings (discussed below).

Statistical inference is often required for all three tasks. For example, one might want to add 95% confidence intervals around descriptive, predictive, or causal estimates involving samples of target populations.

As in most attempts at classification, the boundaries between the above categories are not always sharp. However, this trichotomy provides a useful starting point to discuss the data requirements, assumptions, and analytics necessary to successfully perform each task of data science. A similar taxonomy has been traditionally taught by data scientists from many disciplines, including epidemiology, biostatistics,[12] economics,[13] and political science.[14] Some methodologists have referred to the causal inference task as "explanation",[15] but this is a somewhat misleading term because causal effects may be quantified while remaining unexplained (e.g., randomized trials identify causal effects even if the causal mechanisms that explain them are unknown).

Sciences are primarily defined by their questions rather than by their tools: We define astrophysics as the discipline that learns the composition of the stars, not as the discipline that uses the spectroscope. Similarly, data science is the discipline that describes, predicts, and makes causal inferences (or, more generally, counterfactual predictions), not the discipline that uses machine learning algorithms or other technical tools. Of course, data science certainly benefits from the development of tools for the acquisition, storage, integration, access, and processing of data, as well as from the development of scalable and parallelizable analytics. This data engineering powers the scientific tasks of data science.

**Prediction vs. Causal inference**

Data science has excelled at commercial applications, such as shopping and movie recommendations, credit rating, stock trading algorithms, and advertisement placement. Some of these data scientists have transferred their skills to scientific research with biomedical applications like Google's algorithm to diagnose diabetic retinopathy[16] (after 54 ophthalmologists classified more than 120,000 images), Microsoft's algorithm to predict pancreatic cancer months before its usual diagnosis[17] (using the online search histories of 3000 users who were later diagnosed with cancer), and Facebook's algorithm to detect users who may be suicidal[18] (based on posts and live videos).

All these applications of data science have one thing in common: They are predictive, not causal. They map inputs (e.g., an image of a human retina) to outputs (e.g., a diagnosis of retinopathy), but they do not consider how the world would look like under different courses of action (e.g., would the diagnosis change if we operated on the retina).

Mapping observed inputs to observed outputs is a natural candidate for automated data analysis because this task only requires 1) a large dataset with inputs and outputs, 2) an algorithm that establishes a mapping between inputs and outputs, and 3) a metric to assess the performance of the mapping, often based on a gold standard.[19] Once these three elements are in place, as in the retinopathy example, predictive tasks can then be automated via data-driven analytics that evaluate and iteratively improve the mapping between inputs and outputs without human intervention.

More precisely, the component of prediction tasks that can be easily automated is the one that does not involve any expert knowledge. Prediction tasks require expert knowledge to specify the scientific question—what candidate inputs and what outputs—and to identify/generate relevant data sources.[20] (The extent of expert knowledge varies across different prediction tasks.[21]) However, no expert knowledge is required for prediction after the candidate inputs and the outputs are specified and measured in the population of interest. At this point, a machine learning algorithm can take over the data analysis to deliver a mapping and quantify its performance. The resulting mapping may be opaque, as in many deep learning applications, but its ability to map the inputs to the outputs with a known accuracy in the studied population is not in question.

The role of expert knowledge is the key difference between prediction and causal inference tasks. Causal inference tasks require expert knowledge not only to specify the question (the causal effect of what treatment on what outcome) and to identify/generate relevant data sources, but also to describe the causal structure of the system under study. Causal knowledge, usually in the form of unverifiable assumptions,[22,23] is necessary to guide the data analysis and to provide a justification for endowing the resulting numerical estimates with a causal interpretation. In other words, the validity of causal inferences depends on structural knowledge, which is usually incomplete, to supplement the information in the data. As a consequence, no algorithm can quantify the accuracy of causal inferences from observational data. The following simplified example helps fix ideas about the different role of expert knowledge for prediction versus causal inference.

*Example*

Suppose we want to use a large health records database to predict infant mortality (the output) using clinical and lifestyle factors collected during pregnancy (the inputs). We have just applied our expert knowledge to decide what the outputs and candidate inputs are, and to select a particular database in the population of interest. The only requirement is that the potential inputs need to temporally precede the outputs, regardless of the causal structure linking them. At this point of the process our expert knowledge will not be needed any more: An algorithm can provide a mapping between inputs and outputs at least as good as any mapping we could propose and, in many cases, astoundingly better.

Now suppose we want to use the same health records database to determine the causal effect of maternal smoking during pregnancy on the risk of infant mortality. A key problem is confounding: Pregnant women who do and do not smoke differ in many characteristics (e.g., alcohol consumption, diet, access to adequate prenatal care) that affect the risk of infant mortality. Therefore, a causal analysis needs to identify and adjust for those confounding factors which, by definition, are associated with both maternal smoking and infant mortality. However, not all factors associated with maternal smoking and infant mortality are confounders that should be adjusted for. For example, birthweight is strongly associated with both maternal smoking and infant mortality, but adjustment for birthweight induces bias because birthweight is a risk factor that is itself causally affected by maternal smoking. In fact, adjustment for birth weight results in a bias often referred to as the "birthweight paradox": Low birth weight babies from mothers who smoked during pregnancy have a lower mortality than those from mothers who did not smoke during pregnancy.[24]

An algorithm devoid of causal expert knowledge will rely exclusively on the associations found in the data and is therefore at risk of selecting features, like birthweight, that increase bias. The "birth weight paradox" is indeed an example of how the use of automatic adjustment procedures may lead to the incorrect causal conclusion. In contrast, a human expert can readily identify many variables that, like birthweight, should not be adjusted for because of their position in the causal structure.

Also, a human expert may identify features that should be adjusted for, even if they are not available in the data, and propose sensitivity analyses[25] to assess the reliability of causal inferences in the absence of those features. In contrast, an algorithm that ignores the causal structure will not alert about the need to adjust for features that are not in the data.

Given the central role of (potentially fallible) expert causal knowledge in causal inference, it is not surprising that researchers look for procedures to alleviate the reliance of causal inferences on causal knowledge. Randomization is the best such procedure.

When a treatment is randomly assigned, we can unbiasedly estimate the average causal effect of treatment assignment *in the absence of detailed causal knowledge about the system under study*. Randomized experiments are central in many areas of science in which relatively simple causal questions are asked.[26] Randomized experiments are also commonly used, often under the name A/B testing, to answer simple causal questions in commercial web applications. However, randomized designs are often infeasible, untimely, or unethical in the extremely complex systems studied by health and social scientists.[26]

A failure to grasp the different role of expert knowledge in prediction and causal inference is a common source of confusion in data science (the confusion is compounded by the fact that predictive analytic techniques, such as regression, can also be used for causal inference when combined with causal knowledge).

Both prediction and causal inference require expert knowledge to formulate the scientific question, but only causal inference requires causal expert knowledge to answer the question. As a result, the accuracy of causal estimates cannot be assessed by using metrics computed from the data, even if the data were perfectly measured in the population of interest.

**Implications for decision-making**

A goal of data science is to help make better decisions. For example, in health settings, the goal is to help decision-makers—patients, clinicians, policy-makers, public health officers, regulators—decide among several possible strategies. Frequently, the ability of data science to improve decision making is predicated on the basis of its success at prediction.

However, the premise that predictive algorithms will lead to better decisions is questionable. An algorithm that excels at using data of patients with heart failure to predict who will die within the next five years is agnostic about how to reduce mortality. For example, a prior hospitalization may be identified as a useful predictor of mortality, but nobody would suggest that we stop hospitalizing people in order to reduce mortality. Identifying patients with bad prognosis is very different from identifying the best course of action to prevent or treat a disease. Worse, predictive algorithms, when incorrectly used for causal inference, may lead to incorrect confounder adjustment and therefore conclude, for example, that maternal smoking appears to be beneficial for low birthweight babies.

Predictive algorithms inform us that a decision needs to be made, but they cannot help us make the decision. For example, a predictive algorithm that identifies patients with severe heart failure does not provide information about whether heart transplant is the best treatment option. In contrast, causal analyses are designed to help us make decisions because they tackle "what if" questions. For example, a causal analysis will compare the benefit-risk profile of heart transplant versus medical treatment in patients with certain severity of heart failure.

Interestingly, the distinction between prediction and causal inference (counterfactual prediction) becomes unnecessary for decision making when the relevant expert knowledge can be readily encoded and incorporated into the algorithms. For example, a purely predictive algorithm that learns to play Go can perfectly predict the counterfactual state of the game under different moves, and a predictive algorithm that learns to drive a car can accurately predict the counterfactual state of the car if, say, the brakes are not operated. Because these systems are governed by a set of known game rules (in the case of games like Go) or physical laws with some stochastic components (in the case of engineering applications like self-driving cars), an algorithm can eventually predict the behavior of the entire system under a hypothetical intervention.

Take the game of Go, which has been mastered by an algorithm "without human knowledge".[27] When making a move, the algorithm has access to all information that matters: game rules, current board position, and future outcomes fully determined by the sequence of moves. Further, a reinforcement learning algorithm can collect an arbitrary amount of data by playing more games (that is, by conducting numerous experiments), which allows it to learn by trial-and-error. In this setting, a cleverly designed algorithm running on a powerful computer can spectacularly outperform humans. But this form of causal inference has, at this time in history, a restricted domain of applicability.

Many scientists work on complex systems with partly known and nondeterministic governing laws (the "rules of the game"), with uncertainty about whether all necessary data are available, and for which learning by trial and error—or even conducting a single experiment—is impossible. Even when the laws are known and the data available, the system may still be too chaotic for exact long-term prediction. For example, it was impossible to predict when and where the Chinese space station,[28] while in orbit at an altitude of about 250 km, would fall to Earth.

Consider a causal question about the effect of different epoetin strategies on the mortality of patients with renal disease. We do not understand the causal structure by which molecular, cellular, individual, social, and environmental factors regulate the effect of epoetin dose on mortality risk. As a result, it is currently impossible to construct a predictive model based on electronic health records to reproduce the behavior of the system under a hypothetical intervention on an individual. Some widely publicized disappointments of causal applications of data science, like "Watson for Oncology", have arguably resulted from trying to predict a complex system that is still poorly understood and for which we lack a sound model to combine expert causal knowledge with the available data.[29]

The striking contrast between the cautious attitude of most traditional data scientists (statisticians, epidemiologists, economists, political scientists…) and the "can do" attitude of many computer scientists, informaticians and others seems to be, to a large extent, the consequence of the different complexity of the causal questions historically tackled by each of these groups. Epidemiologists and other data scientists working with extremely complex systems tend to focus on the relatively modest goal of designing observational analyses to answer narrow causal questions about the average causal effect of a variable (i.e., epoetin treatment) as opposed to trying to explain the causal structure of the entire system or to identify globally optimal decision-making strategies.

On the other hand, because newcomers to data science have often focused on systems governed by known laws (like board games or self-driving cars), it is not surprising that they have deemphasized the distinction between prediction and causal inference. Bringing this distinction to the forefront is, however, urgent as an increasing number of data scientists address the causal questions traditionally asked by health and social scientists. Sophisticated prediction algorithms may suffice to develop unbeatable Go software and, eventually, safe self-driving vehicles, but causal inferences in complex systems (say, the effects of clinical strategies to treat a chronic disease) need to rely on data analysis methods equipped with causal knowledge.[30]

**Implications for teaching**

The training of data scientists tends to emphasize the mastery of tools for data management and data analysis. While learning to use these tools will continue to play a central role, it is important that the technical training of data scientists makes clear that the tools are at the service of distinct scientific tasks—description, prediction, and causal inference.

A training program in data science can therefore be explicitly organized in three components, each of them devoted to one of the three tasks of data science. Each component would describe how to articulate scientific questions, data requirements, threats to validity, data analysis techniques, and the role of expert knowledge (separately for description, prediction, and causal inference). This is the approach that we adopted to develop the curriculum of the Clinical Data Science core at the Harvard Medical School, which has now been taught to three cohorts of clinical investigators.

Our students first learn to differentiate between the three tasks of data science, then learn how to generate and analyze data for each task, as well as the differences between tasks. They learn that description and prediction may be affected by selection and measurement biases, but that only causal inference is affected by confounding. After learning predictive algorithms, teams of students compete against each other in a machine learning competition to develop the best predictive model (in an application of the Common Task Framework[2]).

By contrast, after learning causal inference techniques, students understand that a similar competition is not possible because their causal estimates cannot be automatically ranked. Teams with different subject-matter knowledge may produce different causal estimates and often there is no objective way to determine which one is closest to the truth using the existing data.[31]

Then our students learn to ask causal questions in terms of a contrast of interventions conducted over a fixed time period as it would be specified in the protocol of a, possibly hypothetical, experiment, which is the target of inference. For example, to compare the mortality under various epoetin dosing strategies in patients with renal failure, students use subject-matter knowledge to 1) outline the design of the hypothetical randomized experiment that would estimate the causal effect of interest, i.e., the target trial, 2) identify an observational database with sufficient information to approximately emulate the target trial, and 3) emulate the target trial and therefore estimate the causal effect of interest using the observational database. We discuss why causal questions that cannot be translated into target experiments are not sufficiently well defined,[32] and why the accuracy of causal answers cannot be quantified using observational data. In parallel, our students also learn computer coding and the basics of statistical inference to deal with the uncertainty inherent to any data analyses involving description, prediction, or causal inference.

A data science curriculum along the three dimensions of description, prediction, and causal inference facilitates interdisciplinary integration. Learning from data requires paying attention to the different emphases, questions, and analytic methods developed over several decades within statistics, epidemiology, econometrics, computer science, and others. Data scientists without subject-matter knowledge cannot conduct causal analyses in isolation: They don't know how to articulate the questions (what the target experiment is) and they don't know how to answer them (how to emulate the target experiment).

**Conclusion**

Data science is a component of many sciences, including the health and social sciences. Therefore, the tasks of data science are the tasks of those sciences—description, prediction, causal inference. A sometimes overlooked point is that a successful data science requires not only good data and algorithms, but also domain knowledge (including causal knowledge) from its parent sciences.

The current rebirth of data science is an opportunity to rethink data analysis free of the historical constraints imposed by traditional statistics, which has left scientists ill-equipped to handle causal questions. While the clout of statistics in scientific training and publishing impeded the introduction of a unified formal framework for causal inference in data analysis, the coining of the term "data science" and the recent influx of "data scientists" interested in causal analyses provides a one in a generation chance of integrating all scientific questions, including causal ones, within a principled data analysis framework. An integrated data science curriculum can present a coherent conceptual framework that fosters understanding and collaboration between data analysts and domain experts.

On the other hand, if the definitions of data science currently discussed within mainstream statistics take hold, causal inference from observational data will be once more marginalized, leaving health and social scientists on their own. The American Statistical Association statement on *The Role of Statistics in Data Science* (August 8, 2015) makes no reference to causal inference. A recent assessment of data science and statistics[2] did not include the word "causal" (except when mentioning the title of the course "Experiments and Causal Inference"). Heavily influenced by statisticians, many medical editors actively suppress the term "causal" from their publications.[7,33]

A data science that embraces causal inference must (1) develop methods for the integration of sophisticated analytics with expert causal expertise, and (2) acknowledge that, unlike for prediction, the assessment of the validity of causal inferences cannot be exclusively data-driven because the validity of causal inferences also depends on the adequacy of expert causal knowledge. Causal directed acyclic graphs[3,34,35] may play an important role in the development of analytic methods that integrate learning algorithms and subject-matter knowledge. These graphs can be used to represent different sets of causal structures that are compatible with existing causal knowledge and thus to explore the impact of causal uncertainty on the effect estimates.

Large amounts of data could make expert knowledge irrelevant for prediction and for relatively simple causal inferences involving games and some engineering applications. But expert causal knowledge is necessary to formulate and answer causal questions in more complex systems. Affirming causal inference as a legitimate scientific pursuit is the first step to transform data science into a reliable tool to guide decision making.

Finally, the distinction between prediction and causal inference is also crucial to define artificial intelligence (AI). Some data scientists argue that "the essence of intelligence is the ability to predict", and therefore that good predictive algorithms are a form of AI. From this point of view, large chunks of data science can be rebranded as AI (and that is exactly what the tech industry is doing). However, mapping observed inputs to observed outputs barely qualifies as intelligence. Rather, a hallmark of intelligence is the ability to predict *counterfactually* how the world would change under different actions by integrating expert knowledge and mapping algorithms. No AI will be worthy of the name without causal inference.

Table. Examples of tasks conducted by data scientists working with electronic health records.

|  | Data Science Task | | |
|---|---|---|---|
|  | **Description** | **Prediction** | **Causal inference** |
| Example of scientific question | How can women aged 60-80 years with stroke be partitioned in classes defined by their characteristics? | What is the probability of having a stroke next year for women with certain characteristics? | Will starting a statin reduce, on average, the risk of stroke in women with certain characteristics? |
| Data | <ul><li>Eligibility criteria</li><li>Features, e.g., symptoms, clinical parameters…</li></ul> | <ul><li>Eligibility criteria</li><li>Output, e.g., diagnosis of stroke over the next year</li><li>Inputs, e.g., age, blood pressure, history of stroke, diabetes at baseline</li></ul> | <ul><li>Eligibility criteria</li><li>Outcome, e.g., diagnosis of stroke over the next year</li><li>Treatment, e.g., initiation of statins at baseline</li><li>Confounders</li><li>Effect modifiers (optional)</li></ul> |
| Examples of analytics | Cluster analysis … | Regression<br>Decision trees<br>Random forests<br>Support vector machines<br>Neural networks<br>… | Regression<br>Matching<br>Inverse probability weighting<br>G-formula<br>G-estimation<br>Instrumental variable estimation<br>… |